\documentclass[11pt,a4paper]{article}
\pdfoutput=1
\usepackage{authblk}
\usepackage[hyperref]{acl2021}
\usepackage{times,multirow}
\usepackage{latexsym}
\usepackage{xspace}
\newcommand{\libmultilabel}{$\text{\href{https://github.com/ASUS-AICS/LibMultiLabel}{\sf LibMultiLabel}}$\xspace}

\usepackage{enumitem}
\setitemize{noitemsep,topsep=0pt,leftmargin=12pt,parsep=0pt,partopsep=0pt}

\usepackage{microtype}

\aclfinalcopy 
\def\aclpaperid{334} 

\usepackage{amsmath}
\usepackage{caption}
\usepackage{subcaption}

\title{Parameter Selection: Why We Should Pay More Attention to It}

\author[1]{Jie-Jyun Liu}
\author[1]{Tsung-Han Yang}
\author[1,2]{Si-An Chen}
\author[2]{Chih-Jen Lin}
\affil[1]{ASUS Intelligent Cloud Services}
\affil[2]{National Taiwan University}
\affil[ ]{\texttt {\{eleven1\_liu,henry1\_yang\}@asus.com}}
\affil[ ]{\texttt{\{d09922007,cjlin\}@csie.ntu.edu.tw}}

\date{}

\begin{document}
\maketitle
\begin{abstract}
The importance of parameter selection in supervised learning
  is well known. However, due to the many
parameter combinations, an incomplete
or an insufficient procedure is often applied.
This situation may cause misleading or confusing conclusions.
In this opinion paper, through an intriguing example
we point out that the seriousness goes beyond what is generally
recognized. In the topic of multi-label
classification for medical code prediction,
one influential paper conducted a proper parameter selection
on a set, but when moving to a subset of frequently occurring labels,
the authors used the same parameters without a separate tuning.
The set of frequent labels became a popular benchmark
in subsequent studies, which kept pushing the state of the art.
However, we discovered that most of the results in these studies cannot
surpass the approach in the original paper if
a parameter tuning had been conducted at the time.
Thus it is unclear how much progress the subsequent developments have
actually brought.
The lesson clearly indicates that without enough attention
on parameter selection, the research progress in our field
can be uncertain or even illusive.
\end{abstract}

\section{Introduction}
\label{sec:introduction}

The importance of parameter selection in supervised learning
is well known. While parameter tuning has
been a common practice in machine learning and natural language
processing applications, the process remains challenging
due to the huge number of
parameter combinations. The recent trend of applying complicated
neural networks makes the situation more acute.
In many situations,
an incomplete
or an insufficient procedure for parameter selection is applied,
so misleading or confusing conclusions sometimes occur.
In this opinion paper, we present a very intriguing example
showing that,
without enough attention
on parameter selection, the research progress in our field
can be uncertain or even illusive.

In the topic of multi-label
classification for medical code prediction,
\citet{JM18a} is an early work applying deep learning.
The evaluation was
conducted on MIMIC-III and MIMIC-II  \cite{AJ16a},
which may be the most widely used open medical records.
For MIMIC-III, besides using all 8,922 labels, they
follow \citet{HS17a} to check the 50 most frequently occurring labels. We
refer to these two sets respectively as
\begin{center}
  MIMIC-III-full and MIMIC-III-50.
\end{center}
We will specifically investigate MIMIC-III-50.
Based on \citet{JM18a}, many subsequent works made improvements
to push the state of the art.
Examples include
\cite{GW18a,NS18a,XX19a,SCT19a,PC20a,PC20b,SJ20a,FT20a,YC20a,TV20a,HD20a,SJ21a}.

For the data set MIMIC-III-full, \citet{JM18a}
tuned parameters to find the model that achieves the best
validation performance. However, when moving to check
the set MIMIC-III-50, they {\em
  applied the same parameters without a separate tuning.}
We will show that this decision had a profound effect. Many
works directly copied values from \citet{JM18a} for comparison
and presented superior results. However, as demonstrated in
this paper, if parameters for MIMIC-III-50 had been separately
tuned, the approach in \citet{JM18a} easily surpasses most
subsequent developments.
The results fully indicate that parameter selection
is more important than what is generally recognized.

This paper is organized as follows.
In Section \ref{sec:results-past-works},
we analyze past results. The main investigation is in
Section \ref{sec:investigation},
while Section \ref{sec:discussion}
provides some discussion. Some implementation details are in the appendix.
Code and supplementary materials
can be found at \url{http://www.csie.ntu.edu.tw/~cjlin/papers/parameter\_selection}.

\section{Analysis of Works that Compared
with \citet{JM18a}}
\label{sec:results-past-works}

The task considered in \citet{JM18a} is to predict the
associated ICD (International Classification of Diseases) codes
of each medical document. Here an ICD code is referred to as a label.
The neural network considered is 
\begin{equation}
  \label{eq:caml}
  \begin{split}
    & \text{document} \rightarrow \text{word embeddings}\\
    \rightarrow & \text{convolution}
    \rightarrow \text{attention} \rightarrow  \text{linear layer},
  \end{split}
\end{equation}
where the convolutional operation was based on \citet{YK14a}.
A focus in \citet{JM18a} was on the use of attention, so
they detailedly compared the two settings\footnote{After
  convolution, each word is still associated with a short
  vector and attention is a way to obtain a single vector
  for the whole document. For CNN where attention is not used,
  \citet{JM18a} followed \citet{YK14a} to select the maximal
  value across all words.}
\begin{center}
  \begin{tabular}{ll}
    CNN: & \eqref{eq:caml} without attention,
\\
    CAML: & \eqref{eq:caml}.
  \end{tabular}
\end{center}
For the data set MIMIC-III-full,
CAML, which includes an attention layer, was shown to be significantly better
than CNN on all criteria; see Table \ref{tab:camlfull}.
However,
for MIMIC-III-50, the subset of the 50 most frequent labels, the
authors reported in Table \ref{tab:caml50} that CAML is not better
than CNN.

\begin{table}[tb]
  \caption{Experimental results from \citet{JM18a}. Macro-F1 and Micro-F1 are Macro-averaged and Micro-averaged F1 values, respectively.
    P@$n$ is the precision at $n$, the fraction of the $n$
  highest-scored labels that are truly associated with the test instance.}
    \begin{subtable}{0.5\textwidth}
      \centering
  \caption{MIMIC-III-full: 8,922 labels}      
    \begin{tabular}{l|rrr}
      & Macro-F1 & Micro-F1 & P@8 \\ \hline
      CNN &0.042 & 0.419 & 0.581 \\ 
      CAML & 0.088 & 0.539 & 0.709
    \end{tabular}
  \label{tab:camlfull}
\end{subtable}
\begin{subtable}{0.5\textwidth}
  \centering
  \caption{MIMIC-III-50: 50 labels.}
    \begin{tabular}{l|rrr}
      & Macro-F1 & Micro-F1 & P@5 \\ \hline
      CNN &0.576 & 0.625 & 0.620 \\ 
      CAML & 0.532 & 0.614 & 0.609
    \end{tabular}
  \label{tab:caml50}
\end{subtable}
  \label{tab:caml}
\end{table}


\begin{table*}[tb]
      \addtolength{\tabcolsep}{-3pt}  
  \centering
  \caption{MIMIC-III-50 results from past works that have {\em directly  listed
    values} in \citet{JM18a} for comparison. For Macro-F1, please see a note on its definition in the appendix.}
  \begin{tabular}{@{}p{0.37\textwidth}|cccc p{0.26\textwidth}@{}}
    & Macro-F1 & Micro-F1 & P@5 & Code & \multicolumn{1}{c}{Notes} \\ \hline
    \multicolumn{6}{@{}l}{Baseline considered}\\ \hline
        CNN \cite{JM18a}&0.576 & 0.625 & 0.620 & Y &\\
        CAML \cite{JM18a} & 0.532 & 0.614 &  0.609 & Y &
    \\ \hline
    \multicolumn{6}{@{}l}{New network architectures}\\ \hline
    MVC-LDA \cite{NS18a}& 0.597 & 0.668 & 0.644 & N & multi-view convolutional layers\\
    DACNM \cite{PC20b}& 0.579 & 0.641 & 0.616 & N & dilated convolution\\
    BERT-Large \cite{YC20a} & 0.531 & 0.605 & - &N & BERT model\\
    MultiResCNN \cite{FL20a}& 0.606 & 0.670 & 0.641 & Y & multi-filter convolution and residual convolution\\
    DCAN \cite{SJ20a}& 0.615 & 0.671 & 0.642 & Y & dilated convolution, residual connections\\
    G-Coder without additional information \cite{FT20a}  &  - & 0.670 & 0.637 & N & multiple convolutional layers\\
    LAAT \cite{TV20a}& 0.666 & 0.715 & 0.675 & Y & LSTM before attention\\
    \hline
    \multicolumn{6}{@{}p{1\textwidth}}{New network architectures + additional information (e.g.,
    label description, label co-occurrence, label embeddings, knowledge graph, adversarial learning, etc.)}\\ \hline
    LEAM \cite{GW18a} &0.540 & 0.619 & 0.612 & Y & label embeddings used\\
    MVC-RLDA \cite{NS18a}& 0.615 & 0.674 & 0.641 & N & label description used\\
    MSATT-KG \cite{XX19a}& 0.638 & 0.684 & 0.644 & N & knowledge graph\\
    HyperCore \cite{PC20a}& 0.609 & 0.663 & 0.632 & N & label co-occurrence and hierarchy used\\   
    G-Coder with additional information \cite{FT20a}  & - & 0.692 & 0.653 & N & knowledge graph, adversarial learning\\
    GatedCNN-NCI \cite{SJ21a}  & 0.629 & 0.686 & 0.653 & Y & label description used\\   
    \hline   \hline
    \multicolumn{6}{@{}p{1\textwidth}}{Results of our investigation in Section \ref{sec:investigation} are listed below for comparison (values averaged from Table \ref{tab:newresults})}\\ \hline
    CNN  & 0.606 & 0.659 & 0.634 & Y &   \multirow{2}{*}{parameter selection applied} \\ 
    CAML  & 0.635 & 0.684 & 0.651 & Y & \\ 
    \hline
  \end{tabular}
  \label{tab:pastresults}
\end{table*}

The paper \cite{JM18a}
has been highly influential. By exactly using their
training, validation, and test sets for experiments,
many subsequent studies
have
proposed new and better
approaches; see references listed in Section \ref{sec:introduction}.
Most of them
{\em copied}
the CNN and CAML
results from \cite{JM18a} as the baseline for comparison.
Table \ref{tab:pastresults} summarizes
their superior results on MIMIC-III-50.\footnote{
  We exclude papers that used the same MIMIC-III-50 set but did not
  list values in \citet{JM18a} for comparison. Anyway, results in
  these papers are not better than what we obtained in Section \ref{sec:investigation}.
  }
  

\begin{table}[tb]
      \addtolength{\tabcolsep}{-3pt}
\caption{Parameter ranges considered in \citet{JM18a} and the values used.}
\label{tab:parameter_ranges}
\centering
\begin{tabular}{@{} l|p{0.15\textwidth}|rr@{}}
  \multirow{2}{*}{Parameter} &
\multirow{2}{*}{Range} & \multicolumn{2}{c}{Values used}\\ 
&& CNN & CAML \\ \hline  
  $d_c$: \# filters & 50-500 & 500 & 50\\
  $k$: filter size & 2-10 & 4 & 10\\
  $q$: dropout prob. & 0.2-0.8 & 0.2 & 0.2\\
  $\eta$: learning rate & 0.0003, 0.0001, 0.003, 0.001 & 0.003 & 0.0001
\end{tabular}
\end{table}
  
While using the same MIMIC-III-50 set, these
subsequent studies differ from
\citet{JM18a} in various ways.
They proposed sophisticated networks and
may incorporate additional information (e.g., label description, knowledge graph
of words, etc.). Further, they may change settings not considered as parameters
for tuning in \citet{JM18a}. For example,
\citet{JM18a} truncated each document to have at most 2,500 tokens,
but \citet{TV20a} used 4,000.

\section{Investigation}
\label{sec:investigation}
We investigate the performance of the CNN and CAML approaches
in \citet{JM18a} for the set MIMIC-III-50.
Some implementation details are left in supplementary materials.

\subsection{Parameter Selection in \citet{JM18a}}
\label{sec:param-select-citejm1}

\citet{JM18a} conducted parameter tuning on a validation set of
MIMIC-III-full. By considering parameter ranges
shown in Table \ref{tab:parameter_ranges}, 
they applied Bayesian optimization \cite{JS12a}
to choose parameters achieving the highest precision@8 on the
validation set; see the selected values in Table \ref{tab:parameter_ranges}
and the definition of precision in Table \ref{tab:caml}.
However, the following settings are fixed instead of being treated
as parameters for tuning.
\begin{itemize}
\item Each document is truncated to have at most 2,500 tokens.
  Word embeddings are from the CBOW method \cite{TM13a}
  with the embedding size 100.
\item The stochastic gradient method Adam implemented in
  PyTorch is used with its default setting. However, the batch
  size is fixed to be 16 and the learning rate is considered
  as a parameter. Binary cross-entropy loss is
  considered.
\item The Adam method is terminated if the precision@8 does
  not improve for 10 epochs. The model achieving the highest
  validation preision@8 is used to predict the test set for
  obtaining results in Table \ref{tab:camlfull}.
\end{itemize}

Interestingly, for the 50-label
subset of MIMIC-III, \citet{JM18a}
did not conduct a parameter-selection procedure.
Instead, a decision was to
{\em use the same parameters selected for the full-label set}.
Further they switch to present
precision@5 instead of precision@8
because on average each instance is now associated with fewer
labels.

The decision of not separately tuning 
parameters for MIMIC-III-50, as we will see, has a profound effect.
In fact, because in Table \ref{tab:caml50}
CAML is slightly worse than CNN,
\citet{JM18a} have suspected
that a parameter tuning may be needed. They stated that
``we hypothesize
that this\footnote{Here ``this'' means that CAML is not better than
  CNN.} is because the relatively large value of
$k = 10$ for CAML leads to a larger network that is
more suited to larger datasets; tuning CAML's hyperparameters on this dataset would be expected
to improve performance on all metrics.''
However, it seems no subsequent works tried to tune parameters
of CNN or CAML on
MIMIC-III-50.

\begin{table*}[tb]
  \centering
  \caption{MIMIC-III-50 results after parameter selection. We
    consider three random seeds, where 1,337 was used in \citet{JM18a}. Under each seed, we select the five models achieving the best validation precision@5, use them to predict the test set, and report mean/variance.}
  \begin{tabular}{lr|rrrrrrr}
& Seed    & Macro-F1 & Micro-F1 & P@5 \\
\hline
 & 1337 & 0.608 $\pm$ 0.006 & 0.659 $\pm$ 0.005 & 0.634 $\pm$ 0.002 \\ 
CNN  & 1331 & 0.601 $\pm$ 0.013 & 0.660 $\pm$ 0.007 & 0.634 $\pm$ 0.003 \\ 
 & 42 & 0.608 $\pm$ 0.007 & 0.658 $\pm$ 0.006 & 0.633 $\pm$ 0.003 \\ 
\hline
 & 1337 & 0.640 $\pm$ 0.004 & 0.686 $\pm$ 0.004 & 0.650 $\pm$ 0.002 \\ 
CAML  & 1331 & 0.631 $\pm$ 0.004 & 0.682 $\pm$ 0.003 & 0.651 $\pm$ 0.001 \\ 
 & 42 & 0.634 $\pm$ 0.009 & 0.684 $\pm$ 0.004 & 0.651 $\pm$ 0.002 \\ 
\end{tabular}
\label{tab:newresults}
\end{table*}

\subsection{Reproducing Results in \citeauthor{JM18a}}

To ensure the correctness of our implementation, first we reproduce
the results in \citet{JM18a} by considering the following two programs.
\begin{itemize}
\item The public code by \citet{JM18a} at \url{github.com/jamesmullenbach/caml-mimic}.
\item Our implementation of CNN/CAML by following the description in \citet{JM18a}. The code is part of our development on a general multi-label text classification package \libmultilabel.\footnote{\url{https://github.com/ASUS-AICS/LibMultiLabel}}
\end{itemize}
Parameters and
the random seed used in \citet{JM18a} are considered;
see Table \ref{tab:parameter_ranges}.

After some tweaks, on one GPU machine both programs give
{\em exactly the same results} in the following table
\begin{center}
  \begin{tabular}{l|rrr}
    & Macro-F1 & Micro-F1 & P@5 \\ \hline
    CNN & 0.585 & 0.626 & 0.617\\
    CAML & 0.532 & 0.610 & 0.609
  \end{tabular}
\end{center}
Values are very close to those
in Table \ref{tab:caml50}. The small difference might be
due to that our GPUs
or PyTorch versions are not the same as theirs.

We conclude that results in \citet{JM18a} are reproducible.

\subsection{Parameter Selection for MIMIC-III-50}
We apply the parameter-selection
procedure in \citet{JM18a} for MIMIC-III-full to MIMIC-III-50;
see details in Section \ref{sec:param-select-citejm1}.
A difference is that, because training 
MIMIC-III-50 is faster than
MIMIC-III-full, instead of using Bayesian optimization,
we directly check a grid of parameters
that are roughly within the ranges given in Table \ref{tab:parameter_ranges}.
Specifically, we consider
\begin{equation*}
  \begin{split}
    & d_c = 50, 150, 250, 350, 450, 550\\
    & k = 2, 4, 6, 8, 10\\
    & q = 0.2, 0.4, 0.6, 0.8
  \end{split}
\end{equation*}
Because \citet{JM18a} switched to report
test precision@5 for MIMIC-III-50,
for validation we also use
precision@5.

To see the effect of random seeds,
besides the one used in \citet{JM18a},
we checked two other seeds 1,331 and 42, selected solely
because they are the lucky numbers of an author.
  
\subsection{Results and Analysis}

Table \ref{tab:newresults} shows CNN/CAML results after parameter
selection and we have the following observations.
\begin{itemize}
\item Both CNN and CAML achieve better results than those reported
  in Table \ref{tab:caml50} by \citet{JM18a}. The improvement
  of CAML is so significant that it becomes better than
  CNN.
\item From details in supplementary materials, for some parameters
  (e.g., $d_c$ and $q$ for CAML), the selected values are very
  different from those
  used by \citet{JM18a}.
  Thus parameters selected for MIMIC-III-full
  are not transferable to MIMIC-III-50 and a separate tuning is
  essential.
\item Results are not sensitive to
  the random seeds.\footnote{CNN is slight more sensitive to seeds than CAML. More investigation is needed.}
%
\item A comparison with Table \ref{tab:pastresults}
  shows that most subsequent developments cannot surpass
  our CAML results. Some are even inferior to CNN, which is
  the baseline of all these studies.
\item We checked if subsequent developments conducted parameter selection. A summary is in the supplementary materials.
\end{itemize}
Based on our results,
how much progress past works have made is therefore unclear.

\section{Discussion and Conclusions}
\label{sec:discussion}

The intention of this paper is to provide constructive
critiques of past works rather than place blame on their
authors.
For the
many parameter combinations, it is extremely difficult to
check them. However, what our investigation showed is
that if resources or time are available,
more attention should be paid to the parameter selection.
For \citet{JM18a},
as they have done a comprehensive selection on a super-set
MIMIC-III-full,
the same procedure on the simpler MIMIC-III-50 is entirely feasible.
The decision of not doing so leads to a weak baseline in
the subsequent developments.


In conclusion,
besides proposing new techniques such as sophisticated networks,
more attention should be
placed on the parameter selection. In the future this helps to ensure
that strong baselines are utilized to check the progress.


\bibliographystyle{acl_natbib}
\bibliography{sdp}

\appendix
\section{Additional Implementation and Experimental Details}
\label{sec:addit-impl-deta}
Before a stochastic gradient step on a batch of data,
\citet{JM18a} pad sequences with zeros
so that all documents in this batch have the same
number of tokens. Thus results of the forward operation
depend on the batch size. This setting causes issues in validation
because a result independent of the batch size is needed.
Further, for many applications one instance appears
at a time in the prediction stage. Thus we follow \citet{JM18a} to use
\begin{center}
  batch size = 1
\end{center}
in validation and prediction.

Another place that may need a padding operation is in the convolutional
operation. Because a filter sequentially goes over blocks of words, in the
end the output size is smaller than the document length. \citet{JM18a}
did not apply padding for the CNN model, but for the CAML model they
pad each side of the document with zeros so the output has the same
size. We exactly follow their settings.

After the convolutional layer, \citet{JM18a} consider
the $\tanh$ activation function.
For both convolutional and linear layers, a bias term is included.

Before the training process, \citet{JM18a} sort the data according to their lengths. In the stochastic gradient procedure, data are not reshuffled. Therefore, instances considered in each batch are the same across epochs. While this setting is less used in other works, we follow suit to ensure the reproducibility of their results.

In the stochastic gradient procedure, we follow \cite{JM18a} to
set 200 as the maximal number of epochs. This setting is different from
the default 100 epochs in the software \libmultilabel employed for
our experiments. In most situations, the program does not reach
the maximal number of epochs. Instead, it terminates
after the validation P@5 does not improve in 10 epochs.
This criterion also follows from \citet{JM18a}.

All models were trained on one NVIDIA Tesla P40 GPU compatible with the
CUDA 10.2 platform and cuDNN 7.6. Note that results may slightly vary
if experiments are run on different architectures.
 
\section{A Note on Macro-F1}

\citet{JM18a} report macro-F1 defined as 
\begin{center}
      F1 value of macro-precision and macro-recall,
\end{center}
where macro-precision and macro-recall are respectively the mean
of precision and recall over all classes. This definition is
different from the macro-F1 used in most other
works. Specifically, F1 values are obtained for each class first
and their mean is considered as Macro-F1; see the discussion of
the Macro-F1 definitions in \citet{JO21a}. Because works mentioned
in Table \ref{tab:pastresults} may not indicate if they use the
same Macro-F1 formula as \citet{JM18a}, readers should exercise
caution in interpreting Macro-F1 results in
Table \ref{tab:pastresults}. However, based on Micro-F1 and P@5 results
the main point of this paper still stands.

\end{document}